%% file: iclr2015.tex
\documentclass{article} 
\usepackage{iclr2015,times}
\usepackage{url}
\usepackage{graphicx}

\iclrconference 

\title{Recurrent-Neural-Network for Language Detection on Twitter Code-Switching Corpus}

\title{Recurrent-Neural-Network for Language Detection on Twitter Code-Switching Corpus}

\author{
Joseph Chee Chang \& Chu-Cheng Lin  \\
Language Technologies Institute \\
School of Computer Science \\
Carnegie Mellon University \\
Pittsburgh, PA 15213, USA \\
{josephcc, chuchenglin}@cs.cmu.edu
}

\begin{document}

\maketitle

\begin{abstract}
Mixed language data is one of the difficult yet less explored domains of natural language processing. Most research in fields like  machine translation or sentiment analysis assume monolingual input. However, people who are capable of using more than one language often communicate using multiple languages at the same time. Sociolinguists believe this "code-switching" phenomenon to be socially motivated. For example, to express solidarity or to establish authority. Most past work depend on external tools or resources, such as part-of-speech tagging, dictionary look-up, or named-entity recognizers to extract rich features for training machine learning models. In this paper, we train recurrent neural networks with only raw features, and use word embedding to automatically learn meaningful representations. Using the same mixed-language Twitter corpus, our system is able to outperform the best SVM-based systems reported in the EMNLP'14 Code-Switching Workshop by 1\% in accuracy, or by 17\% in error rate reduction.
\end{abstract}

\input{intro}
\input{review}

\input{data}
\input{methodology}

\input{experiments}

\input{conclusion}

\subsubsection*{Acknowledgments}

This work is initiated by the Deep Learning course at the Language Technologies Institute at Carnegie Mellon University offered by Dr. Bhiksha Raj. The authors would also like to thank the teaching assistant Zhenzhong Lan for the insightful discussions.

\bibliography{biblio}
\bibliographystyle{iclr2015}

\end{document}

%% file: intro.tex
\section{Introduction}
\label{sec:intro}

Code--switching refers to the phenomenon that a speaker changes between different languages in a single utterance or conversation. Sociolinguists believe people code--switch because of sociolinguistic motivations, e.g. to express solidarity and familiarity, and to establish authority or distance.\citep{gumperz:82} It is also shown that code--switching has its own grammatical regularities, that switching points almost never occur at certain points.\citep{berkseligson:86} This implies using sophisticated models to identify structures in a mixed-language sentence can potentially help with the language detection task. Also, computational models of code--switching can thus be used to verify linguistic theories. Nonetheless it can also be used to guide downstream NLP applications to use correct language models, which is a practical problem for current social media processing.

There's recently a workshop and shared task on modeling code--switching in social media. The participants were provided code--switched tweets from four language pairs. They then asked to label the tweets for the language \emph{at token level}. This strongly resembles other established sequential labeling problems such as POS tagging and named-entity recognizer (NER); and almost all participants adapted either the conditional random fields model (CRF) or the support vector machine model (SVM). The results are not bad in general. For Nepali--English and Spanish--English most participants were able to get an F1 score around $0.90$ to $0.95$. However, many participants put a lot of efforts into extensive and careful feature engineering. Features used in the shared task include dictionaries, character and word language models, and even output from other NER systems and language identification systems. Despite all the efforts, the outcomes were not always positive. For example, previous work has found that the use of existing NER systems does not translate into improvement of named entities in one submission \citep{king:14}. In comparison, our proposed RNN--based classifier takes only character n--grams, pre--trained embeddings and the texts. Yet our system are competitive against the best results reported in the shared task.

The paper is structured as follows. In Sec.~\ref{sec:related} we review related work in the literature. In Sec.~\ref{sec:data} we describe the code--switched tweets and the annotation scheme in detail. And in Sec.~\ref{sec:method} we describe our RNN--based model. The experiment results are in Sec.~\ref{sec:experiment}.

%% file: review.tex
\section{Related work}
\label{sec:related}

In this section, we will review previous methods for language identification, both in monolingual and multilingual texts, and computational models of code switching. We will also review previous work in neural networks for processing natural language and re-representing words in a semantic vector space.

\paragraph{Monolingual language identification.}
Monolingual language identification is generally treated as a text categorization problem, often defined as given a text $t$, assign its label $l_{t} \in L$ where $L = \{l_1 \ldots l_N\}$ is a predefined set of languages in this task. \citet{baldwin:10} is an excellent review. For the sake of completeness, we include some of the most relevant work here.
By featurizing each text $f(t) \in F$, this problem can be treated as a standard supervised learning problem. Indeed, many well known classifiers such as Naive Bayes\citep{lui:12}, SVM,\citep{jalam:01} and kernel methods\citep{kruengkrai:05} have been used  in the literature. Featurizing each text is an important subproblem. Existing featurization include per--language character frequency, n--grams\citep{cavnar:94}, and linguistically motivated features, such as stop word lists\citep{johnson:93}. People have also achieved excellent accuracy with lower--level features, such as byte sequence\citep{chew:11}.

\paragraph{Multilingual langID.}
The monolinguality of a document is however not always realistic. For example, researchers has found that people have been posting tweets using multiple languages for some time\citep{ling:13}. \citet{lui:14} used topic modeling to model the languages that occur in a document. People have also used sequential labeling algorithms to identify language segments in a document, such as CRF by \citet{king:13}.

\paragraph{Computational models of code--switching.}
Both monolingual and multilingual language identification seek to assign label(s) to a single document or sentence. Computational models of code--switching on the other hand, takes a step further, and tries to pinpoint where the switching between languages happens. Equivalently, one can say code--switching assigns language label at the token level, stead of the sentence level or document level. Linguists have proposed syntactic theories that predicts the locations of code--switching\citep{sankoff:98,belazi:94,cantone:08}, which at the same time prohibits their happening at certain places. Such theories have been incorporated into computational models, such as \citep{li:14}.

\paragraph{Recurrent-Neural-Network for Natural Language Processing}

Due to the sequential natural of natural language data, many problems of natural language processing in the form of tagging are traditionally solved using the linear Hidden Markov Model \citep{kupiec1992robust}. Other frequently used, state-of-the-art machine learning models include Chained-Conditional Random Fields \citep{lafferty2001conditional}, Maximum-Entropy Markov Model \citep{mccallum2000maximum}, and sometimes Support Vector Machines \citep{kudo2001chunking}. More recently, researchers has also begun to explore using RNN architectures to rival, or even outperform such machine learning approaches. \citet{mesnil2013investigation} project 1-hot word vectors using a task-specific, supervised embedding layer, and experimented using both Elman-type \citep{elman1990finding} \citep{mikolov2010recurrent} and Jordan-type \citep{jordan1997serial} RNNs. Their results show that by using the exact same features, RNNs can outperform the state-of-the-art CRF-based model by 1\% F1 score. In our work, we extend their structure to include additional character ngram features, and a pre-trained, generalized embedding layer. Our results show that the proposed architecture trained on only simple lexical features can outperform state-of-the-art SVM-based systems trained on rich features generated using external tools such as named entity recognizers and dictionaries.

%% file: data.tex
\section{Data}
\label{sec:data}
All our experiments are conducted on the Twitter data provided by the EMNLP Code--switching Workshop \citep{solorio2014overview}. Due to Twitter's privacy policy, the organizers were not allowed to provide the tweets themselves. Instead, the participants were provided the unique tweet ids and character offsets. Then, the participants had to crawl the data themselves. With situations like deletion and privacy setting changes made by the users after the initial crawl, we are able to crawl $27,255$ Tweets in total. The language--pair breakdown is listed in Table~\ref{tab:pair-breakdown}.

\begin{table}[ht]
\centering
\begin{tabular}{cc}
\hline
Language pair & number of tweets \\
\hline
English--Spanish & 11,400 \\
English--Nepali & 9,993 \\
Mandarin--English & 995 \\
Modern Standard Arabic--Egyptian Arabic & 5,862 \\
\hline
\end{tabular}
\caption{Number of tweets by language pair.}
\label{tab:pair-breakdown}
\end{table}

Although the main goal of the shared task is to predict the code--switching points, the data is annotated with a finer--grained scheme listed in Table~\ref{tab:scheme}. It should be noted that there is a large variance in each label's frequency. \texttt{mixed} and \texttt{ambiguous}, despite their significance in the linguistic theory of bilinguallism, do not appear to happen often in social media. On the other hand, named entities (labeled with \texttt{ne}) occurs very frequently, to the extent that it was the main deciding factor of the participants' overall performance since \texttt{lang1} and \texttt{lang2} are relatively easy to classify.

\begin{table}[ht]
\centering
\begin{tabular}{cc}
\hline
Label & count \\
\hline
lang1 & 215,014 \\
lang2 & 113,282 \\
ambiguous & 1,536 \\
mixed & 197 \\
other & 72,145 \\
ne & 21,833 \\
\hline
\end{tabular}
\caption{Description of labels and their frequencies.}
\label{tab:scheme}
\end{table}

As stated in Sec.~\ref{sec:intro}, we decide to make the task more realistic (and also harder) by training a classifier that considers all possible languages. Therefore we have a single corpus with (some number) of tweets. The number of tokens by language id is in Table~\ref{tab:lang-breakdown}.

\begin{table}[ht]
\centering
\begin{tabular}{cc}
\hline
Label & count \\
\hline
English (en) & 122,585 \\
Modern Standard Arabic (msa) & 79,484 \\
Nepali (ne) & 60,697 \\
Spanish (es) & 33,099 \\
Egyptian Arabic (arz) & 16,292 \\
\hline
\end{tabular}
\caption{Number of tokens by language id.}
\label{tab:lang-breakdown}
\end{table}

%% file: methodology.tex
\section{Methodology}
\label{sec:method}

The proposed network architecture is based the Elman-type and Jordan-type Recurrent Neural Network. We follow the implementation of Mesnil et. al 2013, and use a 3-word window to capture the immediate context, forming a 7 dimention context vector (1-hot). A supervised word embedding layer is used to project each word onto a 100 dimension real value vector. The second layer is a 100 node hidden layer using the sigmoid function. Finally, we use the softmax function at the output layer. Long-term dependencies are captured using either Elman-type or Jordan-type recurrent structure, where the current output depends on the output of the previous 9 hidden layer or  final layer, respectively.

\begin{figure}
\centering
\includegraphics[width=0.55\textwidth]{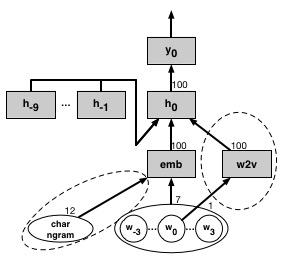}
\caption{Elman-type RNN with Optional Pre-trained Word2Vec and Character Ngram Features.}
\label{fig:arch}
\end{figure}

Extending this architecture to incorporate character ngram features, we use the before mentioned technique to extract a 12-dimension 1-hot vectors from the current word and append it to the 7-dimension context vector, and use the same embedding layer to project them onto the real value vector space. To incorporate the pre-trained Word2Vec model, we use the current word to look up the model, and feed the projected real value vector directly to the hidden layer. Another alternative approach is to replace the supervised embedding layer completely with a pre-trained Word2Vec model, but we think having both a general embedding model and a task specific embedding model can further generalize our method.

In the following subsection, we will talk about the pre-trained Word2Vec model, and also how we extract simple character ngram features. In the evaluation section, we will show the performance of model trained with different network structures and using different feature combinations. 

\subsection{Pre-trained Word2Vector}

We employ skip--gram embeddings as our features. Skip--gram word embeddings \cite{mikolov2013efficient} is a log bilinear model that encourages words with similar contexts to have similar embeddings. We use the implementation from Gensim, and trained on a large Twitter corpus of random samples from the live feed. We randomly sample 10,000 tweets each day, spanning over roughly 2,000 days. We did not filter for specific languages. The idea is that words of different languages tend to share different contexts. So the embeddings should provide good separation between languages. And they proved to provide improvement in the code--switching task \citep{lin:14}.

\subsection{Character NGrams}

Character n--grams prove to be valuable features because languages often have distinct character combinations. For example, a word that starts with ``lle'' is more likely to be Spanish than English. Conversely a word that ends with ``tion'' is more likely to be English. These features are widely used by existing programs, such as \texttt{cld2}\footnote{\url{https://code.google.com/p/cld2/}} and \texttt{ldig}\footnote{\url{https://github.com/shuyo/ldig}}.

In our approach, we extract a fix number of character ngram for each word. We use a window of 3 characters, and extract character bigrams and trigrams from both the begining and end of each word, resulting a 12 dimention character ngram vector. For example, the character ngram vector for the word \emph{architecture} is \emph{[arc, rch, chi, ar, rc, ch, ure, tur, ctu, re, ur, tu]}.

%% file: experiments.tex
\section{Experiments}
\label{sec:experiment}
\subsection{Preliminary Study}

To test the effectiveness of different neural network structures, we first use a smaller data set for a preliminary study. Besides testing both Elman-type and Jordan-type RNN structures, we also tested the effectiveness of our proposed extensions. Our preliminary study dataset contains only a total of 2,734 tweets, where we use 1,000 for training, 1,000 for validation, and the rest for evaluation.

\subsubsection{Experimental Results}

\begin{figure}[ht]
\centering
\includegraphics[width=0.8\textwidth]{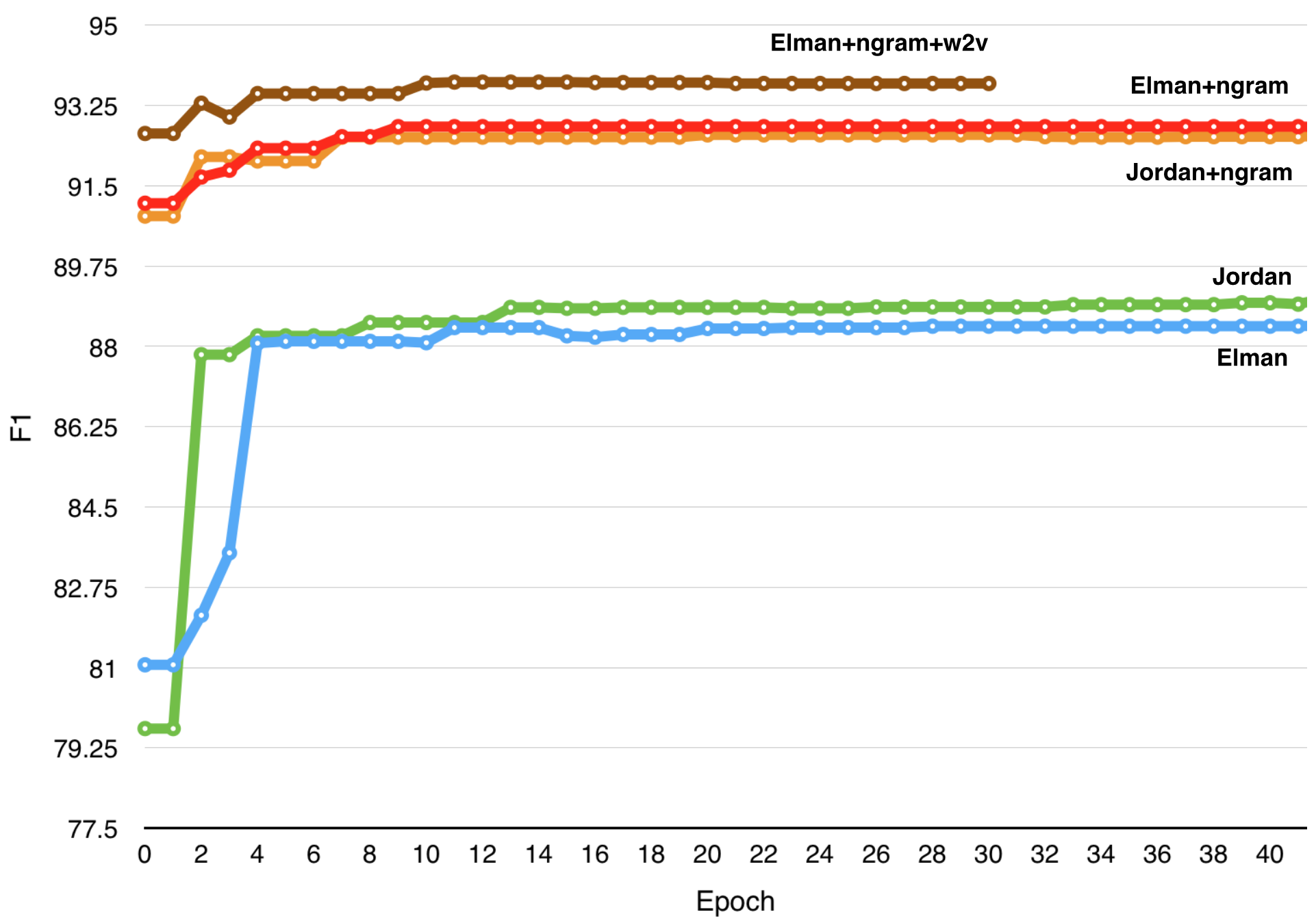}
\caption{Preliminary results of comparing different architecture and feature types, trained on 1,000 tweets.}
\label{fig:prelim}
\end{figure}

In the preliminary study, we tested the five following configurations:

\begin{itemize}
\item \textbf{Jordan}: The basic Jordan-type RNN with input as 7 1-hot word-context vectors and a supervised embedding layer for projecting words onto a real-valued Euclidean space.
\item \textbf{Elman}: The basic Elman-type RNN with input as 7 1-hot word-context vectors and a supervised embedding layer for projecting words onto a real-valued Euclidean space.
\item \textbf{Jordan+ngram}: The Jordan model with an additional 12 1-hot character ngram vector. 
\item \textbf{Elman+ngram}: The Elman model with an additional 12 1-hot character ngram vector.
\item \textbf{Elman+ngram+w2v}: The Elman+ngram configuration with an addtional pre-trained word2vec model feeding directly to the hidden layers.
\end{itemize}

As shown in the figure, with the basic architecture, Jordan-type RNN and Elman-type RNN achieved very similary performance (88.4\% and 88.9\%). Adding character ngrams improves the F1 performance of both architectures by roughly 4\% (92.8\% and 92.6\%). The best performing system that uses Elman architecture and both character ngrams and pre-trained Word2Vec performed a F1 score of 93.7\%. Adding the pre-trained Word2Vec features directly to the hidden layer provided an additional 1.2\% improvement. This suggests that comparing to the basic RNN structures, the proposed two extensions can improve the performance by roughly 5\% on the language identification task.

\subsection{Comparing to Previous Work}

To evaluate our approach, we use the training data of 27,255 tweets obtained from the EMNLP 2014 Code-Switching Workshop. We use 2,734 tweets as our evaluation set, and the rest for training and validation. We acknowledge that we are using a different test set to evaluate our systems, and comparing the results with the systems form the workshop. But we also 
separate the tweets into training, validation, and evaluation sets using disjoint sets of authors. Nevertheless, both the test data we use and the test data from the workshop were collected using the exact same fashion, and both are from none overlapping sets of authors than the training set. 

\subsubsection{Baseline Systems}

The EMNLP Code--Switching Workshop provided a simple deterministic baseline for all categories. Given a word $w$, it looks it up in the training corpus and pick the more frequent language label $\ell(w) \in \{\texttt{lang1}, \texttt{lang2}\}$ as its label. If $\ell(w)=0$ it returns \texttt{other}. And in the case of a tie it returns the language that is more frequent.


\subsubsection{Best Performing Systems in the Shared Task}

According to the overview paper of the workshop \citep{solorio2014overview}, most teams in the shared task adapted the CRF model for language identification. The best Nepali--English submission \cite{barman:14} used SVM as the classification framework. They used character n--grams, context, capitalization, word length, and dictionaries as their features. The best Spanish--English submission \citep{bar:14} also used SVM as the main classifier. They used character n--grams, context, and dictionaries. Additionally they also take language model probabilities as features.


\subsubsection{Experimental Results}

In Table.~\ref{fig:my_label}, we show the performance of our full systems. To compare with the results reported in the workshop, we evaluate our system on the accuracy on two different categories against the best performing systems reported in the EMNLP'14 Code-Switching Workshop. We used only a set of 18,521 tweets to train the networks, because we need the rest of the training set for validation and testing. This is using a significantly smaller training set comparing to previous systems, which have access to the full training data of 27,255 tweets provided by the workshop. 

\begin{figure}[ht]
\centering

\begin{tabular}{r r r}
\hline
 Systems & English-Spanish & English-Nepali \\
 \hline
 Jordan+ngram+w2v & \textbf{95.2}\% & \textbf{96.6}\% \\
 Elman+ngram+w2v & \textbf{95.2}\% & 96.4\% \\
 \hline
 SVM (workshop: dcu-uvt)  & 92.5\% & 96.3\% \\ 
 SVM (workshop: TAU) & 94.2\% & not reported \\
 \hline
 Baseline (LangID) & 75.9\% & 70.0\% \\
 Baseline (Lexical) & 72.6\% & 68.5\% \\
 \hline

\end{tabular}

\caption{Language Identification Accuracy Comparing to the Baseline and Best Reported Systems from the EMNLP'14 Workshop. The parenthesis indicates the size of the training set in number of tweets.}
\label{fig:my_label}
\end{figure}

Similar to the preliminary experiment results, Elman-type and Jordan-type RNNs performed similarly. Comparing to state-of-the-art systems, our networks are able to perform 1\% higher accuracy for English-Spanish category, a 17\% reduction in error rate. For the English-Nepali category, our Jordan-type network performed 0.2\% higher in accuracy, a 8\% reduction in error rate. This shows that the recurrent neural networks and the embedding layers are able to learn meaningful representations for language detection using only the raw lexical features, and are able to rival against SVM-based systems that depend on sophisticated feature extraction to re-represent the input data.

%% file: conclusion.tex
\section{Conclusion}

In this paper, we tackled the important natural language processing task of language detection in a code-switching twitter corpus using recurrent neural networks. This is a novel application for RNN as most previous research focused on using machine learning methods, such as chained conditional random fields, to solve this type of problems. In fact, in the 2014 EMNLP Code-Switching Workshop most of the participating team used CRF to build their models, and the best performing two teams used SVM-based model. Previous work has already compared RNNs to CRF-based model for natural language processing, and results show that RNNs can produce better accuracy by 1\% on a named entity recognition task. We also tested RNN and found that by with the two proposed extensions, RNNs can also outperform state-of-the-art SVM-based system by 1\% in accuracy, or a 17\% in error rate reduction, while using simpler features and smaller training data.